\documentclass[11pt,a4paper]{article}
\usepackage[hyperref]{emnlp2020}
\usepackage{times}
\usepackage{latexsym}

\usepackage{microtype}

\aclfinalcopy 

\usepackage{amsmath}
\usepackage{booktabs}
\usepackage{multirow}
\usepackage{algorithm}
\usepackage{algorithmic}
\usepackage{amssymb}
\usepackage{graphicx}
\usepackage{bbm}
\usepackage{color}
\usepackage{xspace}
\usepackage{xcolor}
\usepackage{capt-of}

\usepackage{url}
\usepackage{listings}
\usepackage{enumitem}
\usepackage{mathtools}
\usepackage{fdsymbol}

\usepackage{todonotes}
\usepackage[normalem]{ulem}
\useunder{\uline}{\ul}{}

\title{Connecting the Dots: A Knowledgeable Path Generator \\for Commonsense Question Answering}

\author{
Peifeng Wang$^{1,3}$,\quad Nanyun Peng$^{1,2,3}$,\quad Filip Ilievski$^{3}$,\quad Pedro Szekely$^{1,3}$,\quad Xiang Ren$^{1,3}$\\
$^{1}$Department of Computer Science, University of Southern California\\
$^{2}$Department of Computer Science, University of California, Los Angeles\\
$^{3}$Information Sciences Institute, University of Southern California
\\
\texttt{\{peifengw,xiangren\}@usc.edu},\quad\texttt{violetpeng@cs.ucla.edu}\\\texttt{\{ilievski,pszekely\}@isi.edu}}

\begin{document}
\maketitle
\begin{abstract}

Commonsense question answering (QA) requires background knowledge which is not explicitly stated in a given context. Prior works use commonsense knowledge graphs (KGs) to obtain this knowledge for reasoning. However, relying entirely on these KGs may not suffice, considering their limited coverage and the contextual dependence of their knowledge. In this paper, we augment a general commonsense QA framework with a \textit{knowledgeable path generator}. By extrapolating over existing paths in a KG with a state-of-the-art language model, our generator learns to connect a pair of entities in text with a dynamic, and potentially novel, multi-hop relational path. Such paths can provide structured evidence for solving commonsense questions without fine-tuning the path generator.
Experiments on two datasets show the superiority of our method over previous works which fully rely on knowledge from KGs (with up to $6\%$ improvement in accuracy), across various amounts of training data. Further evaluation suggests that the generated paths are typically interpretable, novel, and relevant to the task.\footnote{The code is available at \url{https://github.com/wangpf3/Commonsense-Path-Generator}.}

\end{abstract}

\section{Introduction}
\label{sec:intro}

\begin{figure}[t]
  \centering
  \small
  \includegraphics[width=0.9\columnwidth]{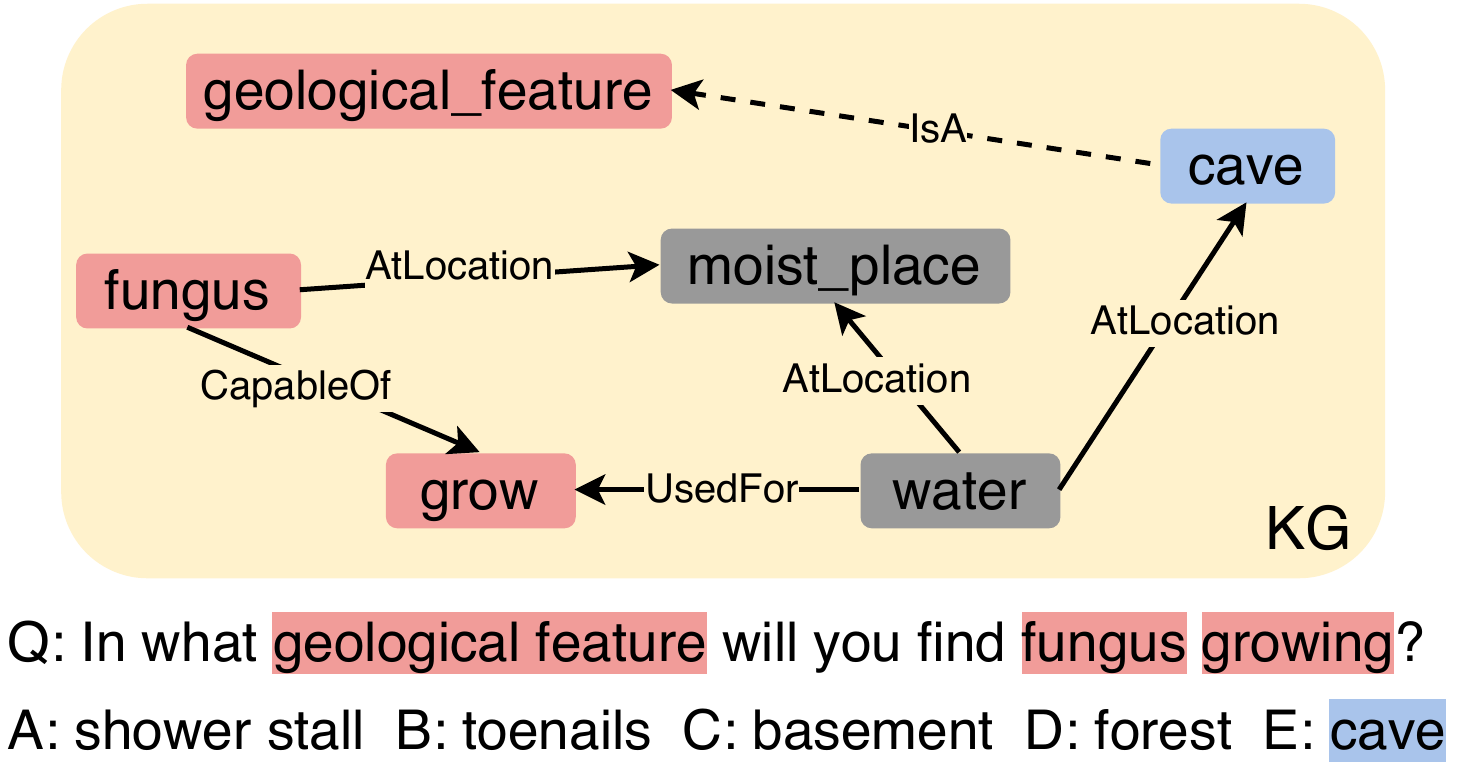}
  \caption{Our path generator learns to connect the question entities (in red) and choice entities (in blue). 
  The dashed arrow indicates a missing link in a static KG.
  }\label{fig:motivation}
\vspace{-1em}
\end{figure}

Solving commonsense QA tasks requires filling gaps with external knowledge.
For instance, given the multiple-choice question in Figure~\ref{fig:motivation}, a system needs to know that \textit{fungus} grows in moist environments, such as \textit{caves}, and that a \textit{cave} is a type of \textit{geological feature}. Such commonsense knowledge is obvious for humans but most existing QA systems do not have it or cannot reason with it. 

Although recent advances in pre-trained language models (LMs) have resulted in impressive performance on commonsense-related benchmarks~\cite{zellers2018swag,bhagavatula2019abductive,huang2019cosmos},
it is unclear whether this is due to commonsense reasoning or to capturing spurious correlations in the data~\cite{niven2019probing}. Pre-trained LMs may answer a question correctly for wrong reasons, making them highly uninterpretable~\citep{mitra2019exploring}. 

Alternatively, a set of systems \textit{retrieve} external knowledge either from large text corpora or knowledge graphs (KGs). 
A corpus, however, might not be an ideal source of commonsense knowledge, as such knowledge is seldom stated explicitly in text~\cite{storks2019commonsense}. In contrast, commonsense KGs, like ConceptNet~\cite{speer2017conceptnet} and ATOMIC~\cite{sap2019atomic}, provide structured evidence about the relevant entities, thus enabling effective reasoning and higher interpretability. Existing systems retrieve knowledge from a KG in the form of: triplets~\cite{mihaylov2018knowledgeable}, multi-hop paths~\cite{lin2019kagnet,bauer2018commonsense}, or subgraphs~\cite{kapanipathi2019infusing}. 

Despite the aforementioned benefits, exploiting these KGs poses the following challenges. Firstly, as KGs are known to suffer from \textit{sparsity}~\cite{li2016commonsense}, 
they might not contain the knowledge needed to fill the gaps between the question and the answer. For example, a missing link \textit{(cave, IsA, geological\_feature)} in Figure~\ref{fig:motivation} might prevent the QA system from choosing the correct answer. Recent work on commonsense KG completion~\cite{li2016commonsense,bosselut2019comet,bosselut2019dynamic} is limited to predicting the tail of a statement with known head and relation, or a single-hop relation between entities. Secondly, due to the large size and heterogeneity of modern KGs, \textit{contextualization}---i.e., identifying a set of KG facts which are relevant or needed to answer a question---is also difficult~\cite{fadnis2019heuristics}. Simply retrieving all paths could introduce noisy information and potentially harm reasoning.


To address this gap between LMs and KGs, we propose a knowledgeable path generator (PG) that generalizes over the facts stored in a KG, rather than only retrieving them. We call our method \textit{neural KG} due to its neural generalization over structured KGs, and, in contrast, we use the term \textit{static KG} for methods which rely exclusively on existing facts in a KG.
Our PG connects a pair of question and answer entities with a (novel) multi-hop path, which may not exist in the KG, allowing for missing facts like \textit{(cave, IsA, geological\_feature)} in Figure~\ref{fig:motivation} to be considered during inference.

To learn such a generator, we: (1) sample a set of random walk instances from a static commonsense KG based on rules and constraints for informativeness and relevance (\textsection\ref{sec:sampling}); 
(2) fine-tune a pre-trained language model --- GPT-2~\cite{radford2019language} on the sampled paths 
(\textsection\ref{sec:ft_gpt}).
By doing so, we transfer the rich knowledge encoded in GPT-2 to our PG. This is expected to both enhance the generalization ability of the PG and combat the sparsity of KGs. Also, by generating high-quality missing links between the question and answer entities, we contextualize the task with relevant commonsense knowledge. 
To understand the impact of our multi-hop PG on downstream commonsense QA tasks, we integrate the PG in an augmented version of a general QA framework (\textsection\ref{sec:qa_sys}).

We run experiments on two benchmark datasets \textit{CommonsenseQA}~\cite{talmor2018commonsenseqa} and \textit{OpenBookQA}~\cite{Mihaylov2018CanAS}. The results show that out method performs better than previous systems augmented with static KGs by up to $\mathbf{6}\%$ in accuracy, which also reveals its potential as a plug-in module for various datasets and as a vital complement to existing KG structures. In the low-resource setting, the accuracy gain over the baselines grows as the  training data decreases, indicating a larger inductive bias of our generator. We also assess the quality and interpretability of our paths through both automatic and human evaluation. 

To summarize, our key contributions are: 
\begin{enumerate}
\vspace{-0.2cm}\item We propose a method to  generate task-relevant knowledge paths that may not exist in the original KG, thus addressing the contextualization and sparsity challenges of KGs.\vspace{-0.2cm}\item We design and implement a framework with three variants of our PG, to understand the role of local and global graph information.  \vspace{-0.2cm}\item Extensive experiments on two benchmark datasets demonstrate the effectiveness of our method compared to previous methods, as well as its robustness to limited training data.
\end{enumerate}


\section{Preliminaries}\label{sec:preliminaries}
\begin{figure}[t]
  \centering
  \includegraphics[width=\columnwidth]{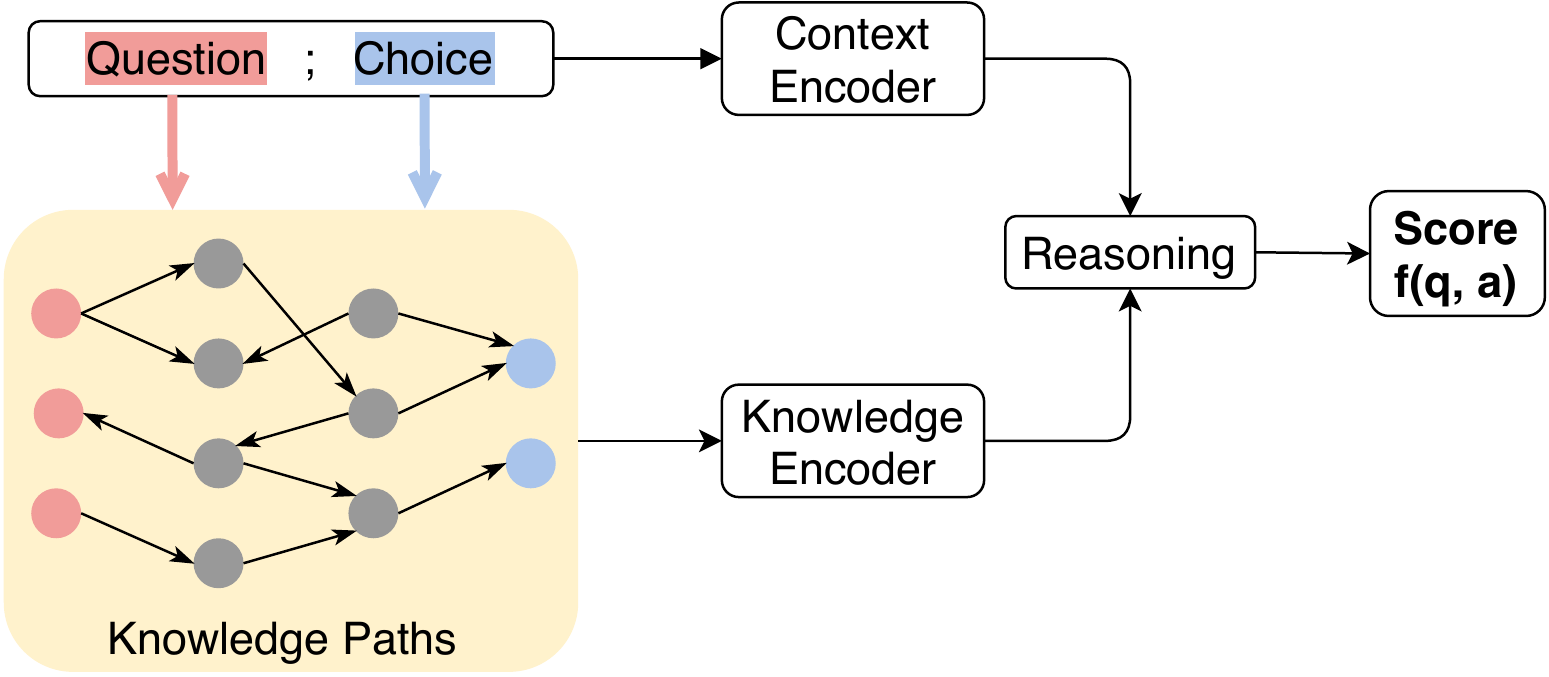}
  \caption{\small \textbf{Our KG-augmented QA Framework}. The reasoning module leverages both the unstructured context and structured knowledge to answer a question.
  }\label{fig:framework}
\vspace{-0.3cm}
\end{figure}

Our multiple-choice commonsense QA setup follows prior work~\cite{talmor2018commonsenseqa,Mihaylov2018CanAS,Bisk2020}: given a question $q$, a system selects exactly one of the choices $a$ as an answer. To experiment with contextualized background knowledge, we adopt a general framework (Figure~\ref{fig:framework}) consisting of a context module, a knowledge module and a reasoning module. The context module encodes both the question $q$ and a choice $a$ as \textit{unstructured evidence}, while the knowledge module encodes external facts as \textit{structured evidence}. Both the unstructured and the structured evidence are fed to the reasoning module, which produces a score for a question-choice pair. The choice with a highest score would be the predicted answer. Next, we introduce each module in detail.

\noindent \textbf{Context Module}
We concatenate a question $q$ and one of its choices $a$ with a special token, and feed the sequence into a contextual encoder. This encoder generates an embedding $\mathbf{c}$, which serves as an unstructured evidence to our system. As commonly done for textual input, we consider a bidirectional pre-trained language model~\cite{devlin2018bert,liu2019roberta} as a contextual encoder. 


\noindent \textbf{Knowledge Module} 
Given a commonsense KG $\mathcal{G}=(\mathcal{E}, \mathcal{R})$, where $\mathcal{E}$ is the entity set and $\mathcal{R}$ is the relation set, we seek a set of relevant knowledge facts for a question-choice pair $\{q,a\}$, which would serve as structured evidence to support reasoning. We employ an entity recognition system to extract relevant entity mentions in the question (denoted by $\mathcal{E}^q=\{e^q\}$) and one of the choices ($\mathcal{E}^a=\{e^{a}\}$). We connect each pair of question-choice entities with a multi-hop path, which can be done either by retrieving existing paths for now (as in previous methods) or by generating paths (see \textsection\ref{sec:qa_sys}). Formally, a path is $p(e^q,e^a)=\{e^q, r_0, e_1, r_1, ..., r_{T-1}, e^a\}$ where $T$ is the number of hops. Note that when $T=1$, the path is a single triplet. 
The set of paths is denoted by $\mathcal{P}=\{p(e^q,e^a)|e^q\in\mathcal{E}^q,e^a\in\mathcal{E}^a\}$.

Naturally, we employ a Relational Network (RN)~\cite{santoro2017simple} to aggregate the retrieved paths into a static knowledge embedding $\mathbf{k}$, which serves as structured evidence. In essence, a RN is a composite function over the set $\mathcal{P}$: 
\begin{equation}
\mathbf{k}=f_\phi(\{g_\theta(p)|p\in\mathcal{P}\}),
\end{equation}
where $f_\phi$ could be any aggregation function and $g_\theta$ could be any neural network which projects a discrete path $p$ into a fixed-size continuous embedding $\mathbf{p}$. We expect that not all paths contribute equally to choosing the right answer. Therefore, we construct the function $f_\phi$ as an attention network:
\begin{equation}\label{eq:phi_agg}
    \mathbf{k}=\sum_{p\in\mathcal{P}} \alpha_p \mathbf{p}.
\end{equation}
We compute the attention weight $\alpha_p$ by using the context embedding $\mathbf{c}$ as a query:
\begin{equation}\label{eq:phi_softmax}
    \alpha_p=\frac{exp(\hat\alpha_p)}{\sum_{p^{'}}\exp{(\hat\alpha_{p^{'}})}},
\end{equation}
where the context embedding $\mathbf{c}$ guides (as an attention query) the encoding of the structured evidence:
\begin{equation}\label{eq:phi_attn}
    \hat\alpha_p=\mathbf{c}^\top\text{tanh}(\mathbf{W}_{att}\cdot\mathbf{p}+\mathbf{b}_{att}).
\end{equation}
Here, the attention network is parameterized by ($\mathbf{W}_{att}$,$\mathbf{b}_{att}$) and tanh($\cdot$) is a nonlinear activation function. Regarding the function $g_\theta$, we employ its original formulation:
\begin{equation}
g_\theta(p)=\text{MLP}[\mathbf{e^q};(\mathbf{r_0}\circ...\circ\mathbf{r_{T-1}});\mathbf{e^a}],
\end{equation}
where $[;]$ is vector concatenation and $\circ$ stands for element-wise multiplication. The components (entities and relations) of a path are represented by their feature vectors.

\noindent \textbf{Reasoning Module}
This module leverages the unstructured evidence (the context embedding $\mathbf{c}$) and the structured one (the knowledge embedding $\mathbf{k}$), to compute the plausibility of a question-choice pair. We concatenate $\mathbf{c}$ with $\mathbf{k}$ and feed them to the final classification layer, which is a linear transformation that scores a question-choice pair $\{q,a\}$:
\begin{equation}\label{eq:cls_layer}
    f(q,a)=\mathbf{W}_{cls}\cdot[\mathbf{c};\mathbf{k}]+\mathbf{b}_{cls},
\end{equation}
The linear classification layer is parameterized by $(\mathbf{W}_{cls},\mathbf{b}_{cls})$. 
We get the final probability over all choices by normalizing with softmax.

\section{Knowledgeable Path Generator}
\begin{figure*}[t]
  \centering
  \includegraphics[width=2\columnwidth]{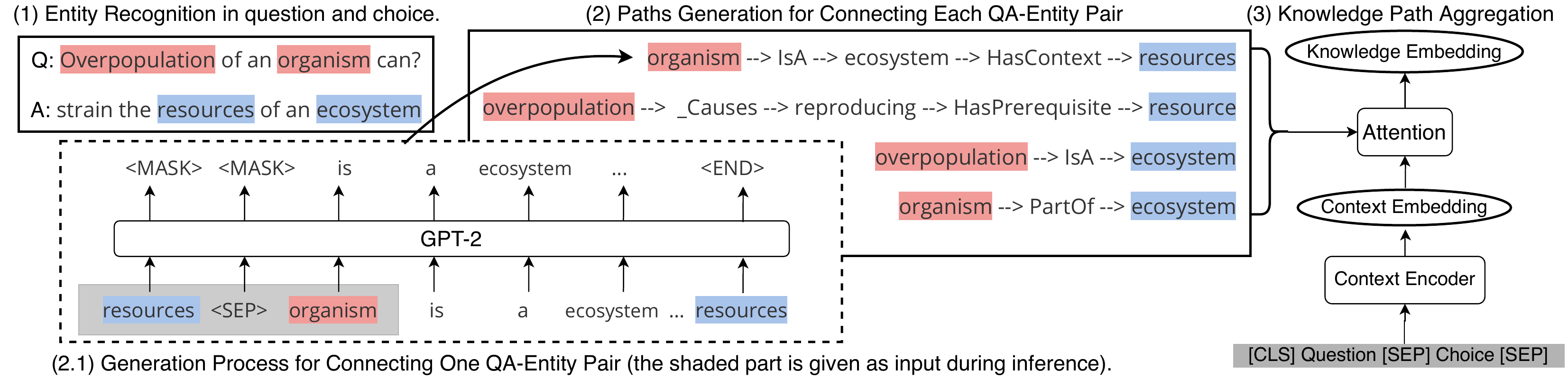}
  \caption{\small \textbf{Overview of our adapted knowledge module.} (1) Extraction of entities from a question and its answer choices. (2) Generation of a multi-hop knowledge path with our PG to connect each pair of question and answer entities. (3) Aggregation of the generated paths into a knowledge embedding.
  }\label{fig:pipeline}
\end{figure*}

Extracting the structured evidence by \textit{retrieving} paths (or subgraphs) from a static KG, as in prior work~\cite{Mihaylov2018CanAS,lin2019kagnet,kapanipathi2019infusing}, faces two key challenges: sparsity and contextualization (\textsection\ref{sec:intro}). We thus propose a knowledgeable path generator (PG), which learns to connect a question-choice entity pair $(e^q,e^a)$ with a multi-hop path. The generated paths are used as structured evidence in the knowledge module. 
Next, we detail the construction of training data (\textsection\ref{sec:sampling}), the learning of our path generator over this data (\textsection\ref{sec:ft_gpt}), and the integration of the generator into the reasoning module (\textsection\ref{sec:qa_sys}). Figure~\ref{fig:pipeline} presents an overview of our adapted knowledge module.

\subsection{Knowledge Path Sampling}\label{sec:sampling}
We sample paths from a commonsense KG using random walks, in order to provide training data for our PG.
Such paths are expected to contain useful knowledge for commonsense QA tasks. 
Given a KG $\mathcal{G}=(\mathcal{E}, \mathcal{R})$, each sampled path $p=\{e_0, r_0, e_1, r_1, ..., r_{T-1}, e_T\}$ is a random walk on the graph, where $e_t\in\mathcal{E}$ and $r_t\in\mathcal{R}$. The number of hops, $T$, is a hyperparameter in our method. 
To improve the quality of the paths, we adopt two heuristic strategies. For \textit{relevance}, we define a subset of relation types that are useful for answering commonsense questions, e.g., \textit{AtLocation} and \textit{IsA}, and filter out the remaining ones, e.g., \textit{RelatedTo}, prior to sampling (see Appendix~\ref{sec:appendix_relation} for the discarded relations). For \textit{informativeness}, we require all relation types in a path to be distinct. 

We explore two sampling strategies in order to select the starting node of the random walks:

\smallskip
\noindent
\textbf{Local Sampling}. The random walks start from the entities that appear in the questions and answer choices of the training set of a benchmark. This strategy is expected to favor generation of paths that are tailored to the task.

\smallskip
\noindent
\textbf{Global Sampling}. We conduct random walks starting from each entity in $\mathcal{E}$. This may divert our PG away from biasing on the local structure of the KG and enhance its generalizability to unseen data.

To include entities that are connected only with inverse triplets in a path, we add a reverse relation $r^{-1}$ for each relation $r$. 
We also sample paths with a mixed number of hops $T$, so our generator can learn to connect entities using paths of variable length, when needed. The full path sampling procedure is described by Algorithm~\ref{algo:random_walk} in the Appendix.

\subsection{Generating Paths to Connect Entities}\label{sec:ft_gpt}

We employ GPT-2~\cite{radford2019language} as the backbone of our path generator. 
GPT-2 is a pre-trained language model that encodes rich unstructured knowledge from large text corpora. 
We foresee two benefits of combining a pre-trained model such as GPT-2 and a static KG: (1) the language model would be able to generate commonsense knowledge paths, by being enriched with relevant \textit{structured} knowledge; (2) the \textit{unstructured} knowledge encoded in the language model would help to alleviate the sparsity challenge of the static KGs.



Unlike COMET~\cite{bosselut2019comet} which fine-tunes GPT (an earlier version of GPT-2) with independent triplets, we fine-tune GPT-2 with consecutive triplets that form paths (see Section~\ref{sec:sampling}). To do so, we first use  GPT-2's Byte-Pair Encoding~\cite{sennrich-etal-2016-neural} to convert each symbolic path $p$ to its textual form 
as a sequence $\{\mathbf{x}_0, \mathbf{y}_0, \mathbf{x}_1, \mathbf{y}_1, ..., \mathbf{y}_{T-1}, \mathbf{x}_T\}$, where $\mathbf{x}_t=\{x_t^1, x_t^2,...,x_t^{\vert e_t\vert}\}$ are phrase tokens of the entity $e_t$ and $\mathbf{y}_t=\{y_t^1, y_t^2, ..., y_t^{\vert r_t\vert}\}$ are phrase tokens of the relation $r_t$. The reverse relations are represented by adding a special prefix token ``\_''. 
The resulting paths mimic natural language sentences to facilitate optimal usage of the knowledge encoded in the pre-trained language model. 
At inference time, in order to connect the question-choice entities, we also 
add the last entity phrase tokens $\mathbf{x}_T$ together with a separate token [SEP] at the beginning of each path sequence, which produces the final transformation $\mathbf{s}^p$.
This informs the generator about the last entity it should output when generating a path.  
Table~\ref{tab:path_samples} provides an example path transformation. 
\begin{table}[t]
\caption{\small \textbf{Example Transformation of a Symbolic Path into Text.}  }\label{tab:path_samples}
	\centering
	\small
	\scalebox{0.9}{
	\begin{tabular}{l}
		\toprule
\{predator, DistinctFrom, prey, IsA, animal\} \\
$\rightarrow$ \{\colorbox{lightgray}{animal, [SEP], predator}, distinct, from, prey, is, a, animal\} \\
		\bottomrule
	\end{tabular}
	}
\end{table}

The PG learns to maximize the probability of the observed paths given the entity pairs. We use negative conditional log likelihood as a loss function:
\begin{equation}
\mathcal{L}=-\sum_{t=\vert \mathbf{x}_0\vert+\vert \mathbf{x}_T\vert+1}^{\vert \mathbf{s}^p\vert}\log P(s_t^p\mid s_{<t}^p),
\end{equation}
where the conditional probability is defined as:
\begin{equation}\label{eq:softmax}
    P(s_t^p\mid s_{<t}^p)=
\text{softmax}(\mathbf{W}_{vocab}\cdot \mathbf{h_t}).
\end{equation}
Here $\mathbf{h_t}$ denotes the final GPT-2 representation for $s_t^p$. $\mathbf{W}_{vocab}$ is the embedding matrix for the token-based vocabulary used by GPT-2, which generalizes well to unseen words.\footnote{This is because an unseen word of an entity or a relation may be split into several tokens that exist in the vocabulary.}
During the inference, the target entity ($e^a$), the [SEP] token, and the starting entity ($e^q)$ are fed to our generator (the shaded part in Table~\ref{tab:path_samples}), and greedy decoding is used to generate a path connecting the two entities. Other constrained decoding strategies would be left as future work.

\subsection{Adapted Commonsense QA Framework}\label{sec:qa_sys}
To facilitate integration of the structured evidence from our path generator instead of a static KG, we adapt the knowledge module from \textsection\ref{sec:preliminaries} slightly. 

We construct the path set $\mathcal{P}$ by generating a multi-hop path $p(e^q,e^a)$ for each pair of a question entity $e^q$ and a choice entity $e^a$ with our PG and greedy decoding. To represent each path with an embedding, we perform mean pooling of the hidden states from the last layer of GPT-2 (before the softmax layer in Eq.~\ref{eq:softmax}) as a new formulation for the function $g_\theta$:
\begin{equation}
    g_\theta(p)=\text{MEAN}(\{\mathbf{h_1,h_2...,h_{\vert s^p\vert}}\}).
\end{equation}
Since GPT-2 has been pre-trained on a large corpus, we believe such representation should be sufﬁcient for preserving the information of the paths. Then, the knowledge embedding obtained with the function $f_\phi$ of the RN (Eq.~\ref{eq:phi_agg}-\ref{eq:phi_attn}) is concatenated with the original static knowledge embedding as our new definition of $\mathbf{k}$.


The whole pipeline is optimized by minimizing its cross-entropy loss. The set of learnable parameters excludes the parameters of our proposed PG, because we observed that fixing their values yields optimal performance. This points to another advantage of our PG: after being fine-tuned on the sampled random walks from a KG, the PG could be integrated within an existing QA system with no further training.

\section{Experiments}
\subsection{Datasets}
We evaluate our method on two commonsense QA benchmarks: \textit{CommonsenseQA}~\cite{talmor2018commonsenseqa} and \textit{OpenBookQA}~\cite{Mihaylov2018CanAS}. 
As the test set of \textit{CommonsenseQA} is not publicly available, the predictions for it can only be evaluated once every two weeks via the official leaderboard. Thus, we report our test score on the leaderboard, and perform more extensive comparisons on the data split used in~\citet{lin2019kagnet}. 
Besides questions and answers, \textit{OpenBookQA} provides a collection of background facts in a textual form. We use the correspondence between these facts and their questions, prepared by~\citet{clark2019f}, as an additional input to the context module for all methods, except RoBERTa-large (see \textsection\ref{sec:results}).


\subsection{KG and Path Data Preparation}
\smallskip
\noindent
\textbf{Entity Recognition}~
We employ ConceptNet~\cite{speer2017conceptnet}, a popular commonsense KG. As stated in \textsection\ref{sec:sampling}, we disregard triplets that belong to a predefined set of relations (see Appendix
). 
Similar to previous work~\cite{lin2019kagnet}, we use lexical matching to ground the entities mentioned in the question and the answer choices to our KG. One exception is that each answer choice in \textit{CommonsenseQA} is treated as a single entity, as these tend to correspond directly to concepts in ConceptNet.

\smallskip
\noindent
\textbf{Path Sampling}
We sample a set of paths with varying lengths, ranging from 1 to 3 hops. 
Global sampling  generates 2,825,692 paths, while local sampling results in 133,612 paths for CommonsenseQA and 105,155 for OpenBookQA.
We split them into training/dev/test sets at a $90:5:5$ ratio.

\subsection{Baselines}
As baselines, we consider a fine-tuned LM, static KG-augmented models, and a 1-hop link predictor on the question and the answer entities.

\smallskip
\noindent
\textbf{Fine-tuned LM}. To examine the role of the external knowledge, we compare to a ``Fine-tuned LM'' ablation of our QA framework without the knowledge module (\textsection\ref{sec:preliminaries}).


\begin{table*}[t]
\caption{\small \textbf{Test accuracy with varying proportions of \textit{CommonsenseQA}} (using the data split in \cite{lin2019kagnet}). Results (as mean and standard deviation) are computed over 4 experimental runs with different random seeds (top score in boldface, second score underlined). Parts of the results for baselines are reported from our another work~\cite{feng2020scalable}.} 
\label{tab:csqa_main}
\centering\small
\scalebox{0.85}{
\begin{tabular}{lcccccc}
\toprule
\multirow{2}{*}{\textbf{Methods}}&
\multicolumn{3}{c}{ \textbf{BERT-large}}&\multicolumn{3}{c}{\textbf{RoBERTa-large}}\\
\cmidrule(lr){2-4} \cmidrule(lr){5-7}
& 20\% Train & 60\% Train & 100\% Train & 20\% Train & 60\% Train & 100\% Train\\
\midrule
 Fine-tuned LM (w/o KG) &  $46.25 ~(\pm0.63)$&  $52.30 ~(\pm0.16)$&  $55.39~(\pm0.40)$&$55.28~(\pm0.35)$&$65.56~(\pm0.76)$&$68.69~(\pm0.56)$\\
\midrule
 + RN& $45.12~(\pm0.69)$ &$54.23~(\pm0.28)$  & $\underline{58.92}~(\pm0.14)$ & $61.32~(\pm0.68)$&$66.16~(\pm0.28)$&$69.59~(\pm3.80)$ \\
 + RGCN&  $48.67~(\pm0.28)$& $54.71~(\pm0.37)$ &$57.13~(\pm0.36)$&$58.58~(\pm0.17)$  & $68.33~(\pm0.85)$ &$68.41~(\pm0.66)$\\
 + GconAttn& $47.95~(\pm0.11)$ & $54.96~(\pm0.69)$ &$56.94~(\pm0.77)$  & $57.53~(\pm0.31)$&$68.09~(\pm0.63)$ &$69.88~(\pm0.47)$\\
 + Link Prediction& $47.10~(\pm0.79)$ & $53.96~(\pm0.56)$ & $56.02~(\pm0.55)$ & $60.84~(\pm1.36)$&$66.29~(\pm0.29)$&$69.33~(\pm0.98)$ \\
\midrule
 + PG-Local& $\underline{50.20}~(\pm0.31)$ &$\underline{55.68}~(\pm0.07)$  & $56.81~(\pm0.73)$ & $61.56~(\pm0.72)$&$67.77~(\pm0.83)$&$70.43~(\pm0.65)$ \\
 + PG-Global& $49.89~(\pm1.03)$ &$55.47~(\pm0.92)$  & $57.21~(\pm0.45)$ & $\underline{62.93}~(\pm0.82)$&$\underline{68.65}~(\pm0.02)$&$\underline{71.55}~(\pm0.99)$\\
 + PG-Full& $\textbf{51.97}~(\pm0.26)$ &$\textbf{57.53}~(\pm0.19)$  & $\textbf{59.07}~(\pm0.30)$ & $\textbf{63.72}~(\pm0.77)$&$\textbf{69.46}~(\pm0.23)$&$\textbf{72.68}~(\pm0.42)$ \\
\bottomrule
\end{tabular}}
\end{table*}

\smallskip
\noindent
\textbf{Static KG Models}.
We compare to three static KG variants of our QA framework that model the knowledge module with path/graph encoders: (1) a RN degenerate version of our system, which computes a knowledge embedding by an attention mechanism over the retrieved paths for each question-choice entity pair; 
(2) Relational Graph Convolutional Networks (RGCN)~\cite{schlichtkrull2018modeling} which encode local graphs by using graph convolutional networks with relation-specific weight matrices; (3) GconAttn~\cite{wang2019improving} which models the alignment between entities via attention and pools over all entity embeddings. 

\smallskip
\noindent
\textbf{Link Prediction Model}. This baseline predicts the relation between question and answer entities instead of creating or finding knowledge paths. Namely, we employ TransE~\cite{bordes2013translating} to learn a representation for every entity and relation in ConceptNet, which is then leveraged to predict a 1-hop relation for each pair of question and answer entities. The representations for each resulting triplet are used as 1-hop path embeddings. The rest of this baseline is identical to our QA framework.


\subsection{Model Variations}
We experiment with three variants of our method which differ in terms of the knowledge embedding: (1) PG-Full:~combination of our global PG and a static RN as detailed in \textsection{\ref{sec:qa_sys}};
(2) PG-Local:~a local PG which is trained on both local and global paths; (3) PG-Global:~a global, data-independent PG which is trained on global paths only. We note that PG-Local and PG-Global do not include the static knowledge embedding.


\begin{table}[tb]
\centering
\caption{\small\textbf{Test accuracy on \textit{OpenBookQA}.} Methods with AristoRoBERTa leverage the textual evidence by~\citet{clark2019f} as an additional input to the context module.}
\label{tab:obqa_kg}
\scalebox{0.68}{
\begin{tabular}{lcc}
\toprule
\textbf{Methods}          & \textbf{RoBERTa-large}     & \textbf{AristoRoBERTa}     \\
\midrule
Fine-tuned LMs (w/o KG)          &  $64.80~(\pm2.37)$   & $78.40~(\pm1.64)$       \\
\midrule
+ RN             & $65.20~(\pm1.18)$    &$75.35~(\pm1.39)$
       \\
+ RGCN           & $62.45~(\pm1.57)$   &$74.60~(\pm2.53)$
       \\
+ GconAtten      & $64.75~(\pm1.48)$   &$71.80~(\pm1.21)$
       \\
+ Link Prediction & $66.30~(\pm0.48)$ & $77.25~(\pm1.11)$         \\
\midrule
+ PG-Local       &  $\underline{70.05}~(\pm1.33)$   &$\underline{79.80}~(\pm1.45)$       \\
+ PG-Global      & $68.40~(\pm0.31)$ & $\mathbf{80.05}~(\pm0.68)$         \\
+ PG-Full        & $\textbf{71.20}~(\pm0.96)$ & $79.15~(\pm0.78)$\\
\bottomrule
\end{tabular}
}
\end{table}
\subsection{Results} \label{sec:results}
\smallskip
\noindent
\textbf{Main Results}
For all systems, 
we experiment with several encoders as a context module: BERT-large~\cite{devlin2018bert} and RoBERTa-large~\cite{liu2019roberta} for \textit{CommonsenseQA}, RoBERTa-large and AristoRoBERTa~\cite{clark2019f} for \textit{OpenBookQA}. Tables~\ref{tab:csqa_main} and \ref{tab:obqa_kg} show the results for CommonsenseQA and OpenBookQA, respectively. 
On both datasets, we observe consistent improvements brought by our method with different context encoders. 
Our full model which, combines both generated and static knowledge, achieves the best performance overall,
suggesting these two knowledge sources are complementary. 
Typically, either our local or global variant yields second best results, demonstrating the effectiveness of the generated paths as structured evidence and their superiority over the static KG methods. The comparable performance of Link Prediction to the static KG methods indicates that even predicting 1-hop knowledge paths helps to address the KG sparsity.

Furthermore, we report comparable results to the other systems on the official test sets, accessible via the leaderboards (Tables~\ref{tab:csqa_sota} and \ref{tab:obqa_sota}). Notably, the two best-performing systems, UnifiedQA~\cite{khashabi2020unifiedqa} and TTTTT~\cite{raffel2019exploring}, are based on the T5 language model~\cite{raffel2019exploring}, which requires excessive computational resources and is impractical in an academic setting. Excluding these, our full method achieves the best performance on both datasets.
\begin{table}[tb]
\centering
\small
\caption{\small\textbf{Test accuracy on \textit{CommonsenseQA}'s official leaderboard}. Note that the SOTA system, UnifiedQA is impractical (11B parameters) in an academic setting.}
\label{tab:csqa_sota}
\scalebox{0.85}{
\begin{tabular}{lcc}
\toprule
\textbf{Methods}& \textbf{Single} & \textbf{Ensemble}           \\
\midrule
RoBERTa~\cite{liu2019roberta} & 72.1 & 72.5 \\

RoBERTa+FreeLB~\cite{zhu2019freelb}&-&73.1\\
RoBERTa+HyKAS~\cite{ma2019towards}&73.2&-\\
XLNet+DREAM&73.3&-\\
RoBERTa+KE &-& 73.3\\
RoBERTa+KEDGN &-& 74.4\\
XLNet+GraphReason~\cite{lv2019graph} & 75.3&-\\
Albert~\cite{lan2019albert} & -&76.5          \\
UnifiedQA\textsuperscript{*}~\cite{khashabi2020unifiedqa} & $\mathbf{79.1}$&-\\
\midrule
Albert+PG-Full& $75.6$ & $\underline{78.2}$ \\
\bottomrule
\end{tabular}
}
\end{table}

\begin{table}[h]
\centering
\caption{\small\textbf{Test accuracy on \textit{OpenBookQA} leaderboard}. All listed methods leverage the provided science facts as additional textual input. Note that the top 2 systems, UnifiedQA (11B parameters) and TTTTT (3B parameters) are computationally expensive and impractical in an academic setting.}
\label{tab:obqa_sota}
\scalebox{0.8}{
\begin{tabular}{lc}
\toprule
Methods         & Test           \\
\midrule
Careful Selection~\cite{banerjee2019careful} & 72.0\\
AristoRoBERTa & 77.8\\
KF + SIR~\cite{banerjee2020knowledge} & 80.0\\
Albert + KB & 81.0\\
TTTTT\textsuperscript{*}~\cite{raffel2019exploring} & $\underline {83.2}$\\
UnifiedQA\textsuperscript{*}~\cite{khashabi2020unifiedqa} &$\textbf{87.2}$          \\
\midrule
AristoRoBERTa + PG-Full      & $80.2$ \\
Albert + PG-Full & 81.8\\
\bottomrule
\end{tabular}
}
\end{table}

\begin{figure}[h]
  \centering
  \includegraphics[width=1\columnwidth]{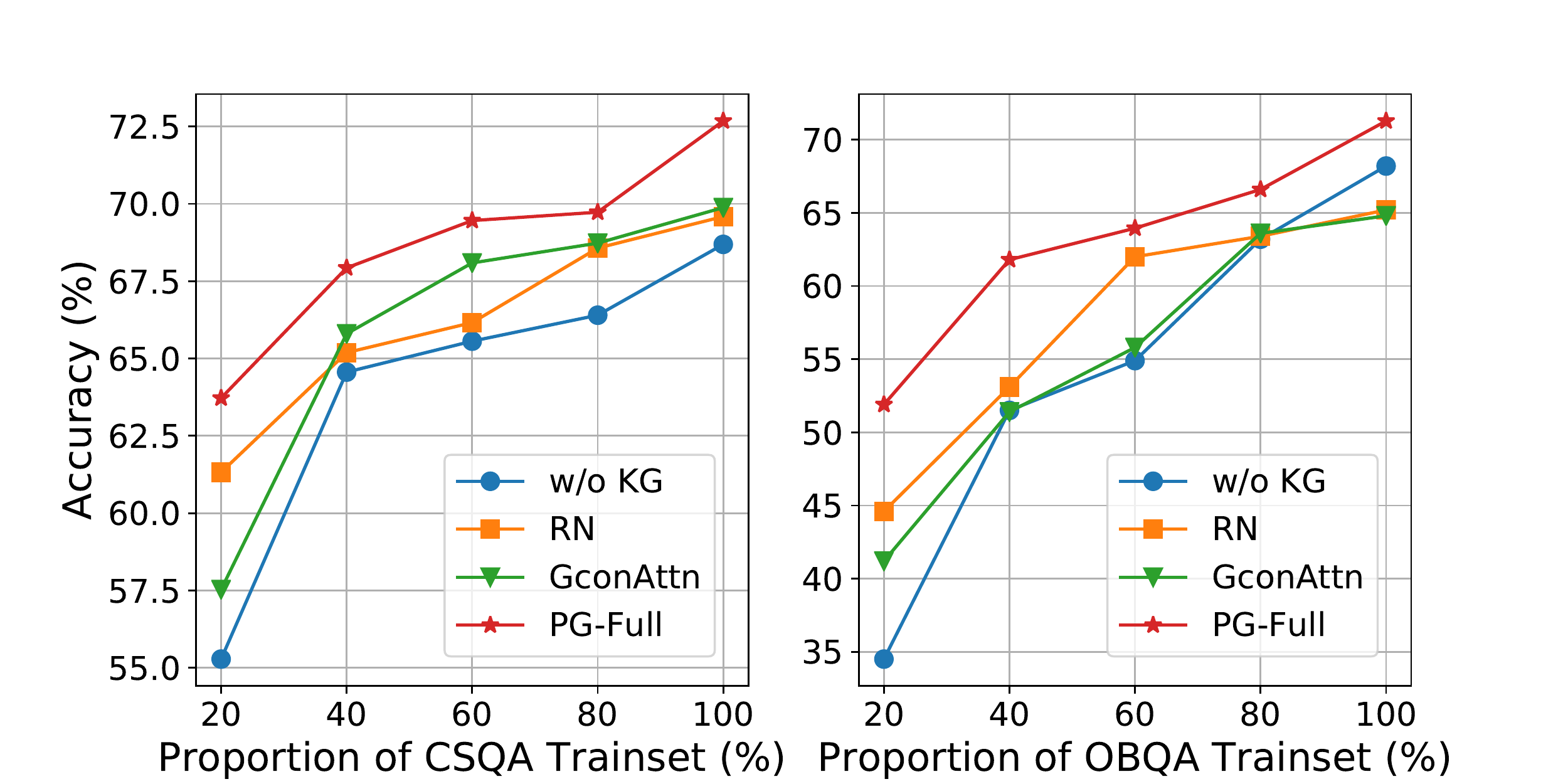}
  \caption{\small Test accuracy on \textit{CommonsenseQA} (left) and \textit{OpenBookQA} (right) with different proportions of training data.}\label{fig:spasity}
\vspace{-0.3cm}
\end{figure}

\smallskip
\noindent
\textbf{Less Labeled Data}
To compare the robustness of our model and the baselines to sparsity, we perform experiments with \{20\%, 40\%, 60\%, 80\%, 100\%\} of the training data from both datasets. The results, displayed in Table~\ref{tab:csqa_main} and Figure~\ref{fig:spasity}, show that our method (with RoBERTa) performs better or equal to the baselines with any amount of training data. The performance gain brought by either our Global or Full model is higher when less data is used, which shows that introducing structured evidence as inductive bias helps in a low-resource setting.

\smallskip
\noindent
\textbf{Ablation Study}
We study the contribution of different strategies for learning our generator based on the performance of our Global and Local variants in Tables~\ref{tab:csqa_main}-\ref{tab:obqa_kg}. We also include another variant by training our path generator from scratch, i.e. training a randomly-initialized model with the same architecture as GPT-2 instead of fine-tuning a pre-trained one. 
This \textit{Scratch} variant achieves $68.75$ and $65.50$ accuracy on the \textit{CommonsenseQA} and \textit{OpenBookQA} test sets, respectively, with RoBERTa-large as the text encoder. Its performance thus resembles that of the static KG baselines while our \textit{Full} method achieves $72.68$ and $71.20$. This demonstrates that learning paths from scratch approximates what a static KG has already, whereas the unstructured knowledge stored in a pre-trained GPT-2 helps to complement missing knowledge in a static KG. When coupled with a more powerful encoder like RoBERTa or Albert, our \textit{Global} variant achieves comparable or better results than our \textit{Local} variant, without fitting the paths to the task, and thus holds a promise to enhance generalization on a wider range of datasets.  

\subsection{Study of Path Quality \& Interpretability}\label{sec:perf_analysis}
\noindent
\textbf{Automatic Evaluation}~~We perform automatic evaluation of the validity and novelty of the generated paths from our \textit{Global} and \textit{Scratch} PG variants.
To automatically measure \textit{validity}, we analyze (1) the proportion of paths which successfully connect the head and the tail entities (\texttt{Connection}), (2) the proportion of entities/relations found in ConceptNet (\texttt{Valid Entity /  Relation}). We also leverage a commonsense knowledge base completion model, \textit{Bilinear AVG}~\cite{li2016commonsense}, which produces a score for a given triplet. This model reportedly achieves $92.5\%$ accuracy on commonsense knowledge completion and has been used in previous work~\cite{bosselut2019comet}. We average the scores of all the triplets in a path which are missing in ConceptNet as its \texttt{Score}. We compute \textit{novelty} as the proportion of paths which contain at least one triplet missing in ConceptNet (\texttt{Novelty}). 
\begin{table}[t]
\centering
\caption{\small \textbf{Automatic and Human Evaluation of the generated Paths on the task testset}. All scores are scaled to be percentage-based.}\label{tab:path_analysis}
\scalebox{0.77}{
\begin{tabular}{lrrrr}
\toprule
\multirow{2}{*}{\textbf{Metric}} & \multicolumn{2}{c}{CommonsenseQA} & \multicolumn{2}{c}{OpenBookQA} \\
\cmidrule(lr){2-3} \cmidrule(lr){4-5}                        & Global          & Scratch          & Global         & Scratch        \\ \midrule
Connection              & 97.33           & 91.16           & 96.03          & 96.01         \\
Valid Entity            & 98.64           & 97.78           & 99.21          & 97.97         \\
Valid Relation          & 100.00          & 100.00          & 100.00         & 100.00        \\
Score                   & 59.31           & 53.27           & 57.74          & 50.62         \\
Novelty                 & 75.82           & 58.18           & 78.93          & 53.81         \\
\midrule
H-Valid                 & 89.20           & 60.13           & 84.93          & 53.73         \\
H-Relevance             & 87.53           & 70.53           & 88.13          & 74.00         \\ 
\bottomrule
\end{tabular}
}
\vspace{-0.2cm}
\end{table}

The results are presented in Table~\ref{tab:path_analysis}. Firstly, our two generator variants are able to connect a vast majority of the entity pairs with a valid path (over $90\%$ \texttt{Connection}). For this purpose, our generators only use the relations in the relation set instead of other, out-of-KG phrases ($100\%$ \texttt{Valid Relation}). In addition, the novel paths from the \textit{Global} generator are of higher quality compared with the ones from the \textit{Scratch} generator, given that any fact with a score over 0.5 is classified as positive by \textit{Bilinear AVG}, which is later confirmed by our human evaluation as well. The \textit{Global} generator also has a higher \texttt{Novelty}, indicating the necessity of transferring knowledge from a pre-trained GPT-2 to complement a static KG. 

\smallskip
\noindent
\textbf{Human Evaluation}~
We also conduct human evaluation on two dimensions of the generated paths: (1) \textit{validity} (How valid are the paths?) (2) \textit{relevance} (How relevant are the paths to the question?). We randomly sample 50 paths from our \textit{Global} and \textit{Scratch} generator for different question-choice entity pairs in the test datasets. For each path, we provide the corresponding question and answer choices as the context. We ask three annotators to score each path from 1 (Not at all) to 5 (Very), resulting in a total of 150 scores for each dimension/generator/dataset. The averages of these scores are reported as \texttt{H-Valid} and \texttt{H-Relevance} in Table~\ref{tab:path_analysis}. For both dimensions, our Global generator achieves higher scores, showing the ability of fine-tuning a pre-trained GPT-2 as our generator to learn the path distribution which is of high quality and relevant to commonsense QA. 

\smallskip
\noindent
\textbf{Path Interpretability.}\label{sec:case_study} 
In Table~\ref{tab:case}, we compare example paths generated by our \textit{Global} and \textit{Scratch} variants to connect the question entities to the gold answer entities. 
In Q1, our \textit{Global} generator provides knowledge about the location of an entity with a 2-hop path, which helps with answering such ``Where'' questions. Although the path from our \textit{Scratch} generator also contains the \textit{AtLocation} relation, its first generated hop (\textit{\_IsA}) is less informative. In Q2, our \textit{Global} generator is able to connect complex ideas about harmony and making peace with a 2-hop path, while the path from the \textit{Scratch} variant contains incorrect information: \textit{peace} is caused by \textit{committing perjury}. In Q3, the path from our \textit{Global} generator is able to predict the relevant property of an entity and realizes that a 1-hop relation suffices in this case. Our \textit{Scratch} variant, however, predicts a less precise relation (\textit{\_HasContext}). 
These cases show the path generalization ability of the fine-tuned pre-trained GPT-2, owed to its unstructured knowledge. 
We refer readers to Table~\ref{tab:case_full} in Appendix for more cases.

\begin{table}[!t]
\footnotesize
\vspace{-0.2cm}
\caption{Paths from question to gold answer entities, with novel and valid triplets in boldface. }\label{tab:case}
	\scalebox{0.73}{
	\begin{tabular}{l}
		\toprule
		\hspace{-0.1in}Q1: Where would you find magazines along side many other printed works? \\
		\hspace{-0.1in}A: doctor. $B^{*}: bookstore$. C: market. D: train station. E: mortuary. \\
		\hspace{-0.1in}PG-Global (2-hop): \{magazine, IsA, book, AtLocation, bookstore\}\\
		\hspace{-0.1in}PG-Scratch: \{magazine, \_IsA, magazine, AtLocation, bookstore\}\\
		\midrule
		\hspace{-0.1in}Q2: If you want harmony, what is something you should try to do with the world? \\
		\hspace{-0.1in}A: take time. B: make noise. C: make war. $D^{*}: make~peace$. E: make haste. \\
		\hspace{-0.1in}PG-Global (2-hop): \{\textbf{harmony, \_MotivatedByGoal, make better world},\\ \hspace{-0.1in}HasPrerequisite,
		make peace\} \\
		\hspace{-0.1in}PG-Scratch: \{harmony, \_UsedFor, committing perjury, Causes, make peace\} \\
		\midrule
		\hspace{-0.1in}Q3: Janet was watching the film because she liked what? \\
		\hspace{-0.1in}A: rejection. B: laughter. $C^{*}: being~entertained$. D: fear. E: bordem. \\
		\hspace{-0.1in}PG-Global (1-hop): \{\textbf{film, \_CausesDesire, being entertained}\}\\
		\hspace{-0.1in}PG-Scratch: \{film, \_HasContext, being entertained\} \\
		\bottomrule
	\end{tabular}
	}
\vspace{-0.1in}
\end{table}

\section{Related Work}
\noindent
\textbf{Multi-hop Reasoning on KGs.}
Recent benchmarks for commonsense QA and related tasks like open domain QA~\cite{yang2018hotpotqa} and reading comprehension~\cite{welbl2018constructing}, require systems to conduct multi-hop reasoning. Existing systems typically employ entity linking to recognize the relevant entities, ground them to a KG, and retrieve the paths from the local graph neighborhood around the entities. The retrieved paths are scored or ranked using graph-based metrics (e,g., PageRank, centrality)~\cite{paul2019ranking,fadnis2019heuristics,bauer2018commonsense}, handcrafted rules~\cite{kapanipathi2019infusing} or neural methods (e.g., attention mechanisms)~\cite{kundu2018exploiting,lin2019kagnet}. Rather than relying on a static KG, our PG is able to generate knowledge paths dynamically, even when these are absent in the KG.

\noindent
\textbf{Dynamic Knowledge Path Generation.}
Several methods generate knowledge paths instead of extracting them from static KGs. ~\citet{asai2019learning} learn reasoning paths by forming sequences of evidence documents, however, their approach relies on the inter-document hyperlinks to establish relations in the constructed KG. The extractor of \citet{fu2019collaborative} retrieves missing facts in order to address the sparsity of KGs. Unlike our work, their setting is limited to knowledge graph completion, where both a query entity and a single query relation are given. The most similar existing work to ours is that by~\citet{bosselut2019dynamic}, which also leverages GPT-2 to dynamically generate knowledge paths. We see two key differences between this method and ours: (1) they expand their paths gradually by predicting the next entity one at a time, while we generate the paths in an end-to-end manner; (2) their method is restricted to a setting where the context could be treated as a single entity and the question - as a query relation, which is not a limitation to our method.

\section{Conclusion}
In this paper, we propose a generator of multi-hop knowledge paths, which provides structured evidence for answering commonsense questions.
The generator, learned by fine-tuning GPT-2 on random walks sampled from ConceptNet, produces a path between each pair of question and answer entities. All generated paths are aggregated into a knowledge embedding and fused with a context embedding given by a text encoder for classification. 
Our QA framework enhanced with this generator outperformes both pre-trained language models and prior KG-augmented methods on two commonsense QA benchmarks. The accuracy gain increases with less training data. 
Furthermore, automatic- and human-based evaluations of the generated paths yield high scores for their validity, novelty, and relevance. Future research should investigate how to optimally fuse the knowledge and the context embeddings. It should also address the ambiguity of the entity mentions in the questions, the answers, and the lexical nodes in ConceptNet.


\section{Acknowledgments}
We thank the anonymous reviewers for their insightful comments. This material is based upon work sponsored by the DARPA MCS program under Contract No. N660011924033 with the United States Office Of Naval Research.

\bibliography{emnlp2020}

\begin{thebibliography}{43}
\expandafter\ifx\csname natexlab\endcsname\relax\def\natexlab#1{#1}\fi

\bibitem[{Asai et~al.(2019)Asai, Hashimoto, Hajishirzi, Socher, and
  Xiong}]{asai2019learning}
Akari Asai, Kazuma Hashimoto, Hannaneh Hajishirzi, Richard Socher, and Caiming
  Xiong. 2019.
\newblock Learning to retrieve reasoning paths over wikipedia graph for
  question answering.
\newblock \emph{arXiv preprint arXiv:1911.10470}.

\bibitem[{Banerjee and Baral(2020)}]{banerjee2020knowledge}
Pratyay Banerjee and Chitta Baral. 2020.
\newblock Knowledge fusion and semantic knowledge ranking for open domain
  question answering.
\newblock \emph{arXiv preprint arXiv:2004.03101}.

\bibitem[{Banerjee et~al.(2019)Banerjee, Pal, Mitra, and
  Baral}]{banerjee2019careful}
Pratyay Banerjee, Kuntal~Kumar Pal, Arindam Mitra, and Chitta Baral. 2019.
\newblock Careful selection of knowledge to solve open book question answering.
\newblock \emph{arXiv preprint arXiv:1907.10738}.

\bibitem[{Bauer et~al.(2018)Bauer, Wang, and Bansal}]{bauer2018commonsense}
Lisa Bauer, Yicheng Wang, and Mohit Bansal. 2018.
\newblock Commonsense for generative multi-hop question answering tasks.
\newblock \emph{arXiv preprint arXiv:1809.06309}.

\bibitem[{Bhagavatula et~al.(2019)Bhagavatula, Bras, Malaviya, Sakaguchi,
  Holtzman, Rashkin, Downey, Yih, and Choi}]{bhagavatula2019abductive}
Chandra Bhagavatula, Ronan~Le Bras, Chaitanya Malaviya, Keisuke Sakaguchi, Ari
  Holtzman, Hannah Rashkin, Doug Downey, Scott Wen-tau Yih, and Yejin Choi.
  2019.
\newblock Abductive commonsense reasoning.
\newblock \emph{arXiv preprint arXiv:1908.05739}.

\bibitem[{Bisk et~al.(2020)Bisk, Zellers, Bras, Gao, and Choi}]{Bisk2020}
Yonatan Bisk, Rowan Zellers, Ronan~Le Bras, Jianfeng Gao, and Yejin Choi. 2020.
\newblock Piqa: Reasoning about physical commonsense in natural language.
\newblock In \emph{Thirty-Fourth AAAI Conference on Artificial Intelligence}.

\bibitem[{Bordes et~al.(2013)Bordes, Usunier, Garcia-Duran, Weston, and
  Yakhnenko}]{bordes2013translating}
Antoine Bordes, Nicolas Usunier, Alberto Garcia-Duran, Jason Weston, and Oksana
  Yakhnenko. 2013.
\newblock Translating embeddings for modeling multi-relational data.
\newblock In \emph{Advances in neural information processing systems}, pages
  2787--2795.

\bibitem[{Bosselut and Choi(2019)}]{bosselut2019dynamic}
Antoine Bosselut and Yejin Choi. 2019.
\newblock Dynamic knowledge graph construction for zero-shot commonsense
  question answering.
\newblock \emph{arXiv preprint arXiv:1911.03876}.

\bibitem[{Bosselut et~al.(2019)Bosselut, Rashkin, Sap, Malaviya, Celikyilmaz,
  and Choi}]{bosselut2019comet}
Antoine Bosselut, Hannah Rashkin, Maarten Sap, Chaitanya Malaviya, Asli
  Celikyilmaz, and Yejin Choi. 2019.
\newblock Comet: Commonsense transformers for automatic knowledge graph
  construction.
\newblock \emph{arXiv preprint arXiv:1906.05317}.

\bibitem[{Clark et~al.(2019)Clark, Etzioni, Khot, Mishra, Richardson,
  Sabharwal, Schoenick, Tafjord, Tandon, Bhakthavatsalam et~al.}]{clark2019f}
Peter Clark, Oren Etzioni, Tushar Khot, Bhavana~Dalvi Mishra, Kyle Richardson,
  Ashish Sabharwal, Carissa Schoenick, Oyvind Tafjord, Niket Tandon, Sumithra
  Bhakthavatsalam, et~al. 2019.
\newblock From'f'to'a'on the ny regents science exams: An overview of the
  aristo project.
\newblock \emph{arXiv preprint arXiv:1909.01958}.

\bibitem[{Devlin et~al.(2018)Devlin, Chang, Lee, and
  Toutanova}]{devlin2018bert}
Jacob Devlin, Ming-Wei Chang, Kenton Lee, and Kristina Toutanova. 2018.
\newblock Bert: Pre-training of deep bidirectional transformers for language
  understanding.
\newblock \emph{arXiv preprint arXiv:1810.04805}.

\bibitem[{Fadnis et~al.(2019)Fadnis, Talamadupula, Kapanipathi, Ishfaq, Roukos,
  and Fokoue}]{fadnis2019heuristics}
Kshitij Fadnis, Kartik Talamadupula, Pavan Kapanipathi, Haque Ishfaq, Salim
  Roukos, and Achille Fokoue. 2019.
\newblock Heuristics for interpretable knowledge graph contextualization.
\newblock \emph{arXiv preprint arXiv:1911.02085}.

\bibitem[{Feng et~al.(2020)Feng, Chen, Lin, Wang, Yan, and
  Ren}]{feng2020scalable}
Yanlin Feng, Xinyue Chen, Bill~Yuchen Lin, Peifeng Wang, Jun Yan, and Xiang
  Ren. 2020.
\newblock Scalable multi-hop relational reasoning for knowledge-aware question
  answering.
\newblock \emph{arXiv preprint arXiv:2005.00646}.

\bibitem[{Fu et~al.(2019)Fu, Chen, Qu, Jin, and Ren}]{fu2019collaborative}
Cong Fu, Tong Chen, Meng Qu, Woojeong Jin, and Xiang Ren. 2019.
\newblock Collaborative policy learning for open knowledge graph reasoning.
\newblock \emph{arXiv preprint arXiv:1909.00230}.

\bibitem[{Huang et~al.(2019)Huang, Bras, Bhagavatula, and
  Choi}]{huang2019cosmos}
Lifu Huang, Ronan~Le Bras, Chandra Bhagavatula, and Yejin Choi. 2019.
\newblock Cosmos qa: Machine reading comprehension with contextual commonsense
  reasoning.
\newblock \emph{arXiv preprint arXiv:1909.00277}.

\bibitem[{Kapanipathi et~al.(2019)Kapanipathi, Thost, Patel, Whitehead,
  Abdelaziz, Balakrishnan, Chang, Fadnis, Gunasekara, Makni
  et~al.}]{kapanipathi2019infusing}
Pavan Kapanipathi, Veronika Thost, Siva~Sankalp Patel, Spencer Whitehead,
  Ibrahim Abdelaziz, Avinash Balakrishnan, Maria Chang, Kshitij Fadnis, Chulaka
  Gunasekara, Bassem Makni, et~al. 2019.
\newblock Infusing knowledge into the textual entailment task using graph
  convolutional networks.
\newblock \emph{arXiv preprint arXiv:1911.02060}.

\bibitem[{Khashabi et~al.(2020)Khashabi, Khot, Sabharwal, Tafjord, Clark, and
  Hajishirzi}]{khashabi2020unifiedqa}
Daniel Khashabi, Tushar Khot, Ashish Sabharwal, Oyvind Tafjord, Peter Clark,
  and Hannaneh Hajishirzi. 2020.
\newblock Unifiedqa: Crossing format boundaries with a single qa system.
\newblock \emph{arXiv preprint arXiv:2005.00700}.

\bibitem[{Kundu et~al.(2018)Kundu, Khot, Sabharwal, and
  Clark}]{kundu2018exploiting}
Souvik Kundu, Tushar Khot, Ashish Sabharwal, and Peter Clark. 2018.
\newblock Exploiting explicit paths for multi-hop reading comprehension.
\newblock \emph{arXiv preprint arXiv:1811.01127}.

\bibitem[{Lan et~al.(2019)Lan, Chen, Goodman, Gimpel, Sharma, and
  Soricut}]{lan2019albert}
Zhenzhong Lan, Mingda Chen, Sebastian Goodman, Kevin Gimpel, Piyush Sharma, and
  Radu Soricut. 2019.
\newblock Albert: A lite bert for self-supervised learning of language
  representations.
\newblock \emph{arXiv preprint arXiv:1909.11942}.

\bibitem[{Li et~al.(2016)Li, Taheri, Tu, and Gimpel}]{li2016commonsense}
Xiang Li, Aynaz Taheri, Lifu Tu, and Kevin Gimpel. 2016.
\newblock Commonsense knowledge base completion.
\newblock In \emph{Proceedings of the 54th Annual Meeting of the Association
  for Computational Linguistics (Volume 1: Long Papers)}, pages 1445--1455.

\bibitem[{Lin et~al.(2019)Lin, Chen, Chen, and Ren}]{lin2019kagnet}
Bill~Yuchen Lin, Xinyue Chen, Jamin Chen, and Xiang Ren. 2019.
\newblock Kagnet: Knowledge-aware graph networks for commonsense reasoning.
\newblock \emph{arXiv preprint arXiv:1909.02151}.

\bibitem[{Liu et~al.(2019)Liu, Ott, Goyal, Du, Joshi, Chen, Levy, Lewis,
  Zettlemoyer, and Stoyanov}]{liu2019roberta}
Yinhan Liu, Myle Ott, Naman Goyal, Jingfei Du, Mandar Joshi, Danqi Chen, Omer
  Levy, Mike Lewis, Luke Zettlemoyer, and Veselin Stoyanov. 2019.
\newblock Roberta: A robustly optimized bert pretraining approach.
\newblock \emph{arXiv preprint arXiv:1907.11692}.

\bibitem[{Lv et~al.(2019)Lv, Guo, Xu, Tang, Duan, Gong, Shou, Jiang, Cao, and
  Hu}]{lv2019graph}
Shangwen Lv, Daya Guo, Jingjing Xu, Duyu Tang, Nan Duan, Ming Gong, Linjun
  Shou, Daxin Jiang, Guihong Cao, and Songlin Hu. 2019.
\newblock Graph-based reasoning over heterogeneous external knowledge for
  commonsense question answering.
\newblock \emph{arXiv preprint arXiv:1909.05311}.

\bibitem[{Ma et~al.(2019)Ma, Francis, Lu, Nyberg, and
  Oltramari}]{ma2019towards}
Kaixin Ma, Jonathan Francis, Quanyang Lu, Eric Nyberg, and Alessandro
  Oltramari. 2019.
\newblock Towards generalizable neuro-symbolic systems for commonsense question
  answering.
\newblock \emph{arXiv preprint arXiv:1910.14087}.

\bibitem[{Mihaylov et~al.(2018)Mihaylov, Clark, Khot, and
  Sabharwal}]{Mihaylov2018CanAS}
Todor Mihaylov, Peter Clark, Tushar Khot, and Ashish Sabharwal. 2018.
\newblock Can a suit of armor conduct electricity? a new dataset for open book
  question answering.
\newblock In \emph{EMNLP}.

\bibitem[{Mihaylov and Frank(2018)}]{mihaylov2018knowledgeable}
Todor Mihaylov and Anette Frank. 2018.
\newblock Knowledgeable reader: Enhancing cloze-style reading comprehension
  with external commonsense knowledge.
\newblock \emph{arXiv preprint arXiv:1805.07858}.

\bibitem[{Mitra et~al.(2019)Mitra, Banerjee, Pal, Mishra, and
  Baral}]{mitra2019exploring}
Arindam Mitra, Pratyay Banerjee, Kuntal~Kumar Pal, Swaroop Mishra, and Chitta
  Baral. 2019.
\newblock Exploring ways to incorporate additional knowledge to improve natural
  language commonsense question answering.
\newblock \emph{arXiv preprint arXiv:1909.08855}.

\bibitem[{Niven and Kao(2019)}]{niven2019probing}
Timothy Niven and Hung-Yu Kao. 2019.
\newblock Probing neural network comprehension of natural language arguments.
\newblock \emph{arXiv preprint arXiv:1907.07355}.

\bibitem[{Paul and Frank(2019)}]{paul2019ranking}
Debjit Paul and Anette Frank. 2019.
\newblock Ranking and selecting multi-hop knowledge paths to better predict
  human needs.
\newblock \emph{arXiv preprint arXiv:1904.00676}.

\bibitem[{Radford et~al.(2019)Radford, Wu, Child, Luan, Amodei, and
  Sutskever}]{radford2019language}
Alec Radford, Jeffrey Wu, Rewon Child, David Luan, Dario Amodei, and Ilya
  Sutskever. 2019.
\newblock Language models are unsupervised multitask learners.
\newblock \emph{OpenAI Blog}, 1(8).

\bibitem[{Raffel et~al.(2019)Raffel, Shazeer, Roberts, Lee, Narang, Matena,
  Zhou, Li, and Liu}]{raffel2019exploring}
Colin Raffel, Noam Shazeer, Adam Roberts, Katherine Lee, Sharan Narang, Michael
  Matena, Yanqi Zhou, Wei Li, and Peter~J Liu. 2019.
\newblock Exploring the limits of transfer learning with a unified text-to-text
  transformer.
\newblock \emph{arXiv preprint arXiv:1910.10683}.

\bibitem[{Santoro et~al.(2017)Santoro, Raposo, Barrett, Malinowski, Pascanu,
  Battaglia, and Lillicrap}]{santoro2017simple}
Adam Santoro, David Raposo, David~G Barrett, Mateusz Malinowski, Razvan
  Pascanu, Peter Battaglia, and Timothy Lillicrap. 2017.
\newblock A simple neural network module for relational reasoning.
\newblock In \emph{Advances in neural information processing systems}, pages
  4967--4976.

\bibitem[{Sap et~al.(2019)Sap, Le~Bras, Allaway, Bhagavatula, Lourie, Rashkin,
  Roof, Smith, and Choi}]{sap2019atomic}
Maarten Sap, Ronan Le~Bras, Emily Allaway, Chandra Bhagavatula, Nicholas
  Lourie, Hannah Rashkin, Brendan Roof, Noah~A Smith, and Yejin Choi. 2019.
\newblock Atomic: an atlas of machine commonsense for if-then reasoning.
\newblock In \emph{Proceedings of the AAAI Conference on Artificial
  Intelligence}, volume~33, pages 3027--3035.

\bibitem[{Schlichtkrull et~al.(2018)Schlichtkrull, Kipf, Bloem, Van Den~Berg,
  Titov, and Welling}]{schlichtkrull2018modeling}
Michael Schlichtkrull, Thomas~N Kipf, Peter Bloem, Rianne Van Den~Berg, Ivan
  Titov, and Max Welling. 2018.
\newblock Modeling relational data with graph convolutional networks.
\newblock In \emph{European Semantic Web Conference}, pages 593--607. Springer.

\bibitem[{Sennrich et~al.(2016)Sennrich, Haddow, and
  Birch}]{sennrich-etal-2016-neural}
Rico Sennrich, Barry Haddow, and Alexandra Birch. 2016.
\newblock \href {https://doi.org/10.18653/v1/P16-1162} {Neural machine
  translation of rare words with subword units}.
\newblock In \emph{Proceedings of the 54th Annual Meeting of the Association
  for Computational Linguistics (Volume 1: Long Papers)}, pages 1715--1725,
  Berlin, Germany. Association for Computational Linguistics.

\bibitem[{Speer et~al.(2017)Speer, Chin, and Havasi}]{speer2017conceptnet}
Robert Speer, Joshua Chin, and Catherine Havasi. 2017.
\newblock Conceptnet 5.5: An open multilingual graph of general knowledge.
\newblock In \emph{Thirty-First AAAI Conference on Artificial Intelligence}.

\bibitem[{Storks et~al.(2019)Storks, Gao, and Chai}]{storks2019commonsense}
Shane Storks, Qiaozi Gao, and Joyce~Y Chai. 2019.
\newblock Commonsense reasoning for natural language understanding: A survey of
  benchmarks, resources, and approaches.
\newblock \emph{arXiv preprint arXiv:1904.01172}.

\bibitem[{Talmor et~al.(2018)Talmor, Herzig, Lourie, and
  Berant}]{talmor2018commonsenseqa}
Alon Talmor, Jonathan Herzig, Nicholas Lourie, and Jonathan Berant. 2018.
\newblock Commonsenseqa: A question answering challenge targeting commonsense
  knowledge.
\newblock \emph{arXiv preprint arXiv:1811.00937}.

\bibitem[{Wang et~al.(2019)Wang, Kapanipathi, Musa, Yu, Talamadupula,
  Abdelaziz, Chang, Fokoue, Makni, Mattei et~al.}]{wang2019improving}
Xiaoyan Wang, Pavan Kapanipathi, Ryan Musa, Mo~Yu, Kartik Talamadupula, Ibrahim
  Abdelaziz, Maria Chang, Achille Fokoue, Bassem Makni, Nicholas Mattei, et~al.
  2019.
\newblock Improving natural language inference using external knowledge in the
  science questions domain.
\newblock In \emph{Proceedings of the AAAI Conference on Artificial
  Intelligence}, volume~33, pages 7208--7215.

\bibitem[{Welbl et~al.(2018)Welbl, Stenetorp, and
  Riedel}]{welbl2018constructing}
Johannes Welbl, Pontus Stenetorp, and Sebastian Riedel. 2018.
\newblock Constructing datasets for multi-hop reading comprehension across
  documents.
\newblock \emph{Transactions of the Association for Computational Linguistics},
  6:287--302.

\bibitem[{Yang et~al.(2018)Yang, Qi, Zhang, Bengio, Cohen, Salakhutdinov, and
  Manning}]{yang2018hotpotqa}
Zhilin Yang, Peng Qi, Saizheng Zhang, Yoshua Bengio, William~W Cohen, Ruslan
  Salakhutdinov, and Christopher~D Manning. 2018.
\newblock Hotpotqa: A dataset for diverse, explainable multi-hop question
  answering.
\newblock \emph{arXiv preprint arXiv:1809.09600}.

\bibitem[{Zellers et~al.(2018)Zellers, Bisk, Schwartz, and
  Choi}]{zellers2018swag}
Rowan Zellers, Yonatan Bisk, Roy Schwartz, and Yejin Choi. 2018.
\newblock Swag: A large-scale adversarial dataset for grounded commonsense
  inference.
\newblock \emph{arXiv preprint arXiv:1808.05326}.

\bibitem[{Zhu et~al.(2019)Zhu, Cheng, Gan, Sun, Goldstein, and
  Liu}]{zhu2019freelb}
Chen Zhu, Yu~Cheng, Zhe Gan, Siqi Sun, Thomas Goldstein, and Jingjing Liu.
  2019.
\newblock Freelb: Enhanced adversarial training for language understanding.
\newblock \emph{arXiv preprint arXiv:1909.11764}.

\end{thebibliography}
\bibliographystyle{acl_natbib}
\appendix

\section{Algorithm for Path Sampling}
\begin{algorithm}[h]\small
	\caption{Path Sampling}
	\label{algo:random_walk}
	\renewcommand{\algorithmicrequire}{\textbf{Input:}}
	\renewcommand{\algorithmicensure}{\textbf{Output:}}
	\begin{algorithmic}[1]
	\REQUIRE $\mathcal{G}=(\mathcal{E}, \mathcal{R})$ and a set of all the question entities $\{e^q\}$
	\ENSURE A set of triplet paths $\{p\}$.
	\REPEAT
	\IF{Do Global Sampling}
	\STATE current\_node $u\leftarrow uniform\_sample(\mathcal{E})$
	\ELSE 
	\STATE current\_node $u\leftarrow~uniform\_sample (\{e^q\})$
	\ENDIF
	\STATE $p\leftarrow\{u\}$
	\FOR{$t=1$ to $T$}
	\STATE $N\leftarrow Neighbor(u)$
	\STATE next\_node $v\leftarrow uniform\_sample(N)$
	\STATE $M\leftarrow All\_Relations(u, v)$
	\WHILE{TRUE}
	\STATE  $r\leftarrow uniform\_sample(M)$
	\IF{$r$ not in $p$}
	\STATE BREAK
	\ENDIF
	\ENDWHILE
	\STATE $p\leftarrow p\cup\{r, v\}$
	\STATE $u\leftarrow v$
	\ENDFOR
	\UNTIL{Maximum number of paths achieved.}
	\end{algorithmic}
\end{algorithm}


\section{Discarded Relations}\label{sec:appendix_relation}
When sampling knowledge paths, we discard some relation types which are regarded to be uninformative and offer little help for answering the questions. They include
\emph{RelatedTo}, \emph{Synonym}, \emph{Antonym}, \emph{DerivedFrom}, \emph{FormOf}, \emph{EtymologicallyDerivedFrom} and
\emph{EtymologicallyRelatedTo}.

\begin{table}[h]
\centering
\small
\caption{QA Dataset Statistics.}
\label{tab:dataset_stat}
\begin{tabular}{lrrr}
\toprule        & Train & Dev   & Test  \\
\midrule
CommonsenseQA (official)    & 9,741 & 1,221 & 1,140 \\
CommonsenseQA (\citeauthor{lin2019kagnet}) & 8,500 & 1,221 & 1,241 \\
OpenBookQA                  & 4,957 & 500   & 500  \\
\bottomrule
\end{tabular}
\end{table}
\section{Datasets Split}
Both CommonsenseQA\footnote{\url{https://www.tau-nlp.org/commonsenseqa}} and OpenbookQA\footnote{\url{https://leaderboard.allenai.org/open_book_qa/submissions/public}} have their datasets available on their leaderboard pages. The dataset split used in~\cite{lin2019kagnet} is also available by request and we have included it as a supplementary material.

\begin{table}[h]
\centering
\small
\caption{Learning rate of different context modules for \textit{CommonsenseQA}.}
\label{tab:hp_csqa}
\begin{tabular}{lrr}
\toprule
                  & Learning Rate & Batch Size \\
\midrule
BERT-large        & 2e-5          & 32         \\
RoBERTa-large     & 2e-6          & 16         \\
Albert-xxlarge-v2 & 1e-5          & 16   \\ 
\bottomrule
\end{tabular}
\end{table}
\begin{table}[h]
\centering
\small
\caption{Learning rate of different context modules for \textit{OpenBookQA}.}
\label{tab:hp_obqa}
\begin{tabular}{lrr}
\toprule
                  & Learning Rate & Batch Size \\
\midrule
Roberta-large     & 1e-5          & 32         \\
AristoRoBERTa     & 2e-5          & 16         \\
Albert-xxlarge-v2 & 1e-5          & 16 \\ 
\bottomrule
\end{tabular}
\end{table}

\section{Implementation Details}
\smallskip
\noindent
\textbf{Path Generator Training} We employ a pre-trained GPT2-base model~\cite{radford2019language} to initialize our generator. Then we fine-tune the generator with an initial learning rate of $1e-5$ and a batch size of $64$. The learning rate is changed with a warm-up period of $500$ mini batches and then linearly decayed. The training lasts until the loss on the development set no longer decreases for 2 epochs. 

\smallskip
\noindent
\textbf{Training on the Task Datasets}
We search for the optimal hyper-parameters based on the classification accuracy on the development set. The learning rate for the context module is chosen from $\{2e-6, 5e-6, 1e-5,2e-5,5e-5\}$. The learning rate for the rest of the parameters is set to $1e-3$. The batch size is chosen from $\{8, 16, 32, 64, 128\}$. A large batch size is achieved by accumulating gradient through several small batches. The training lasts until the accuracy on the development set no longer increases for 2 epochs. The optimal hyper-parameters for both datasets are listed in Tables~\ref{tab:hp_csqa}-\ref{tab:hp_obqa}.

\begin{table*}[h]
\small
\centering
\caption{Number of parameters of the major modules in our QA framework.}
\label{tab:model_size}
\begin{tabular}{lr}
\toprule               & \# Parameters \\
\midrule
BERT-large        & 340M          \\
RoBERTa-large     & 355M          \\
AristorRoBERTa    & 355M          \\
Albert-xxlarge-v2 & 223M          \\
GPT2-base         & 117M          \\
RN                & 399K\\
\bottomrule
\end{tabular}
\end{table*}
\smallskip
\noindent
\textbf{Model Size} We list the model size of the major modules in our QA framework in Table~\ref{tab:model_size}. These include the different pre-trained LMs used as a context module, the backbone of our PG (GPT-2), and the RN used for the static knowledge module.

\begin{table*}[h]
\footnotesize
\caption{More Paths from questions to gold answer entities, with novel and valid triplets in boldface.}
\label{tab:case_full}
	\centering
	\small
	\begin{tabular}{l}
		\toprule
Q1: He spent all summer in his room playing video games, because of this it wasn't surprising\\ for Mother to find a stack of dirty dishes in her what?\\
$A^*$: son's room. B: party. C: dishwasher. D: restaurant kitchen. E: shoes\\
PG-Global: \{play\_video, \_UsedFor, \textbf{computer, AtLocation, son's room}\}\\
PG-Scratch: \{play\_video, \_UsedFor, machine, \_IsA, son's room\}\\
\midrule
Q2: What do people typically do while playing guitar?\\
A: cry. B: hear sounds. $C^*$: singing. D: arthritis. E: making music. \\
PG-Global: \{guitar, Usedfor, \textbf{playing music, HasSubevent, singing}\} \\
PG-Scracth: \{guitar, HasContext, music, \_Causes, singing\}\\
\midrule
Q3: Blue read material outside of his comfort zone because he wanted to gain what?\\
$A^*$: new perspective. B: entertained. C: understanding. D: hunger. E: tired eyes. \\
PG-Global: \{\textbf{reading material, \_HasPrerequisite, learning about subject, Causes, new perspective}\}\\
PG-Scratch: \{reading material, \_HasSubevent, reading, Causes, new perspective\}\\
\midrule
Q4: Bob the lizard lives in a warm place with lots of water.  Where does he probably live?\\
A: rock. $B^*$: tropical rainforest. C: jazz club. D: new mexico. E: rocky places.\\
PG-Global: \{\textbf{warm place, \_AtLocation, forest}, \_IsA, tropical rainforest\}\\
PG-Scracth: \{warm place, \_AtLocation, tropical rainforest\}\\
\midrule
Q5: She was always helping at the senior center, it brought her what?\\
A: satisfaction. B: heart. C: feel better. D: pay. E: happiness.\\
PG-Global: \{help, \_UsedFor, giving assistance, Causes, happiness\} \\
PG-Scratch: \{help, \_HasSubevent, giving assistance, MotivatedByGoal, happiness\}\\
\midrule
Q6: What is likely to satisfy someone's curiosity?\\
$A^*$: hear news. B: read book. C: see favorite show. D: comedy show. E: go somewhere. \\
PG-Global: \{curiosity, CausesDesire, find information, HasSubevent, read, \_Hasprerequisite, hear news\}\\
PG-Scratch: \{curiosity, CausesDesire, hear news\}\\
\midrule
Q7: Where would a person be doing when having to wait their turn?\\
A: have patience. B: get in line. C: sing. $D^*$: stand in line. E: turn left.\\
PG-Global: \{wait, \_HasPrerequisite, stand in line\}\\
PG-Scratch: \{wait, \_HasPrerequisite, stand in line\}\\
\midrule
Q8: It's easier for human's to survive in:\\
A: a cave. B: the ocean. $C^*$: a town. D: alone.\\
PG-Global: \{survive              \_MotivatedByGoal, \textbf{live, \_UsedFor, townhouse, AtLocation, town}\}\\
PG-Scratch: \{survive, \_HasProperty, town\}\\
\midrule
Q9: A man wanted to find the United States on a visual, where should he look?\\
A: history book.  $B^*$: atlas.  C: tv channels.  D: northern hemisphere.  E: map.\\
PG-Global: \{\textbf{visual, \_HasContext, map}, AtLocation, atlas\}\\
PG-Scratch: \{visual, \_IsA, atlas\}\\
\midrule
Q10: What leads to someone going to to bed?\\
A: bad dreams.  B: lazyness.  C: get pregnant.  $D^*$: sleepiness.  E: rest.\\
PG-Global: \{bed, UsedFor, sleeping, Causes, sleepiness\}\\
PG-Scratch: \{bed, UsedFor, sleepiness\}\\
		\bottomrule
	\end{tabular}
\end{table*}
\end{document}